\documentclass{article}

\usepackage{microtype}
\usepackage{graphicx}
\usepackage{subcaption}
\usepackage{cuted}
\usepackage{afterpage}
\usepackage{booktabs} % for professional tables
\usepackage{amsmath,amsfonts,bm}

\def\eqref#1{equation~\ref{#1}}
\def\1{\bm{1}}

\DeclareMathAlphabet{\mathsfit}{\encodingdefault}{\sfdefault}{m}{sl}
\SetMathAlphabet{\mathsfit}{bold}{\encodingdefault}{\sfdefault}{bx}{n}

\usepackage[colorlinks,
            linkcolor=red,
            anchorcolor=red,
            citecolor=violet
            ]{hyperref}      % hyperlinks
\usepackage{url}
\usepackage{booktabs,multirow,makecell}
\usepackage[table]{xcolor}
\definecolor{rowhl}{RGB}{233,241,250}
\newcommand{\best}[1]{\textbf{#1}}
\usepackage{graphicx}
\usepackage{subcaption}   
\usepackage{adjustbox}
\usepackage{wrapfig}
\usepackage{caption} % 更好看的 caption
\usepackage{pgffor} % 用于循环
\usepackage{hyperref}
\newcommand{\blue}[1]{\textcolor{black}{#1}}

\usepackage[preprint]{icml2026}

\usepackage{amsmath}
\usepackage{amssymb}
\usepackage{mathtools}
\usepackage{amsthm}

\usepackage[capitalize,noabbrev]{cleveref}

\theoremstyle{plain}

\theoremstyle{definition}

\theoremstyle{remark}

\usepackage[textsize=tiny]{todonotes}

\icmltitlerunning{Controlling Your Image via Simplified Vector Graphics}

\begin{document}

\twocolumn[
  \icmltitle{Controlling Your Image via Simplified Vector Graphics}
  % It is OKAY to include author information, even for blind submissions: the
  % style file will automatically remove it for you unless you've provided
  % the [accepted] option to the icml2026 package.

  % List of affiliations: The first argument should be a (short) identifier you
  % will use later to specify author affiliations Academic affiliations
  % should list Department, University, City, Region, Country Industry
  % affiliations should list Company, City, Region, Country

  % You can specify symbols, otherwise they are numbered in order. Ideally, you
  % should not use this facility. Affiliations will be numbered in order of
  % appearance and this is the preferred way.
  
  \icmlsetsymbol{equal}{*}
  \icmlsetsymbol{corresponse}{$\dagger$}

  \begin{icmlauthorlist}
    \icmlauthor{Lanqing Guo}{equal,ut}
    \icmlauthor{Xi Liu}{equal,cu}
    \icmlauthor{Yufei Wang}{corresponse,spr}
    \icmlauthor{Zhihao Li}{spr}
    \icmlauthor{Siyu Huang}{corresponse,cu}
    %\icmlauthor{}{sch}
    %\icmlauthor{}{sch}
  \end{icmlauthorlist}

  \icmlaffiliation{ut}{The University of Texas at Austin, USA}
  \icmlaffiliation{cu}{Clemson University, USA}
  \icmlaffiliation{spr}{SparcAI Inc., USA}

  \icmlcorrespondingauthor{Yufei Wang}{yufei.wang@sparclab.ai}
  \icmlcorrespondingauthor{Siyu Huang}{siyuh@clemson.edu}

  % You may provide any keywords that you find helpful for describing your
  % paper; these are used to populate the "keywords" metadata in the PDF but
  % will not be shown in the document
  \icmlkeywords{Machine Learning, ICML}
  \vskip 0.3in

  {%
\renewcommand\twocolumn[1][]{#1}%
\begin{center}
    \centering
    \vspace{-5pt}
    \captionsetup{type=figure}
    \includegraphics[width=1.0\linewidth]{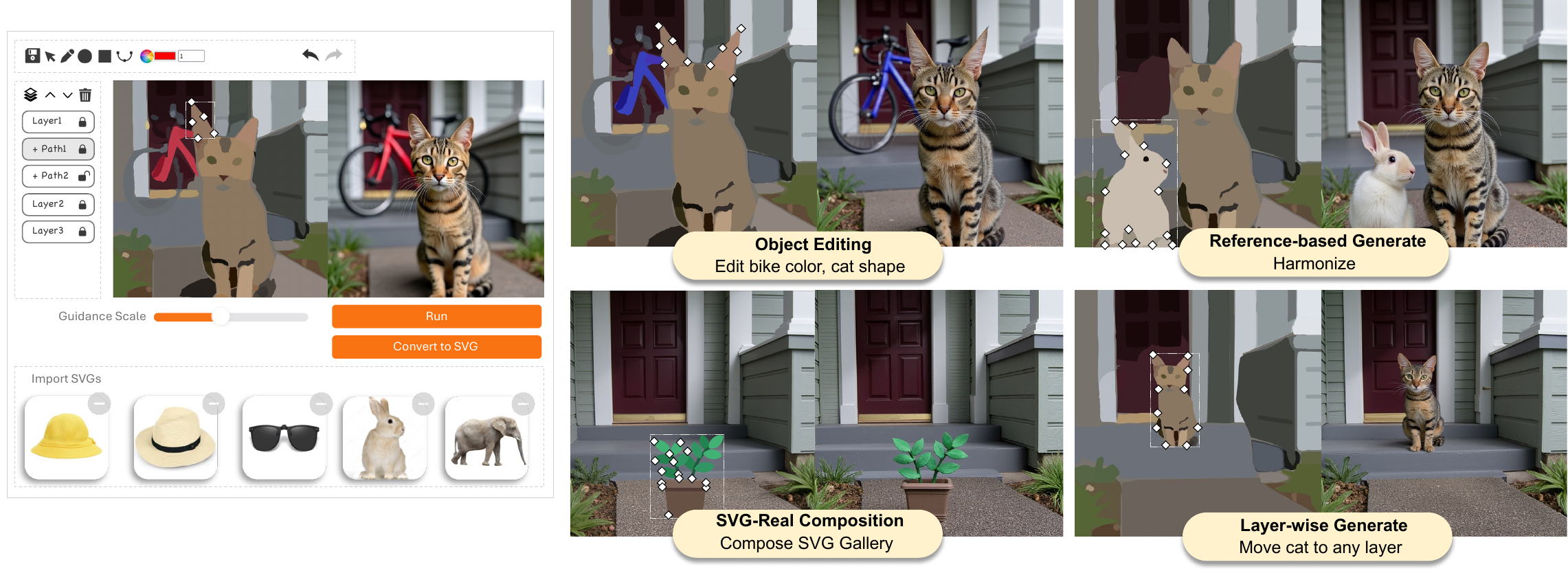}
    \vspace{-10pt}
     \caption{Visual examples of applications of the proposed \textbf{Vec2Pix} framework, where the left column depicts input SVGs and the right column presents the corresponding generated images.
  Vec2Pix offers
1) \emph{Easy-to-control}: it supports layer-wise object insertion, removal, modification, color adjustment, shape editing, and flexible composition; 
2) \emph{High fidelity}: it caches semantic and color information through hierarchical SVG representations; 
and 3) \emph{Strong input-generation alignment}: it ensures precise semantic and structure alignment between SVG inputs and generated images. 
}
    \label{fig:teaser}
    \vspace{5pt}
    \centering
\end{center}%
}
]

% \begin{figure*}[h]
%   \centering
%   % \vspace{-12pt}
%   \includegraphics[width=\textwidth]{fig/teaser1.pdf}  % 或 .png/.jpg
%   \vspace{-16pt}
%   \caption{Visual examples of applications of the proposed \textbf{Vec2Pix} framework, where the left column depicts input SVGs and the right column presents the corresponding generated images.
%   Vec2Pix offers
% 1) \emph{Easy-to-control}: it supports layer-wise object insertion, removal, modification, color adjustment, shape editing, and flexible composition; 
% 2) \emph{High fidelity}: it caches semantic and color information through hierarchical SVG representations; 
% and 3) \emph{Strong input-generation alignment}: it ensures precise semantic and structure alignment between SVG inputs and generated images. 
% }
%   \label{fig:teaser}
% \end{figure*}

% this must go after the closing bracket ] following \twocolumn[ ...

% This command actually creates the footnote in the first column listing the
% affiliations and the copyright notice. The command takes one argument, which
% is text to display at the start of the footnote. The \icmlEqualContribution
% command is standard text for equal contribution. Remove it (just {}) if you
% do not need this facility.

% Use ONE of the following lines. DO NOT remove the command.
% If you have no special notice, KEEP empty braces:
\printAffiliationsAndNotice{}  % no special notice (required even if empty)
% Or, if applicable, use the standard equal contribution text:
% \printAffiliationsAndNotice{\icmlEqualContribution}

\begin{abstract}
Recent advances in image generation have achieved remarkable visual quality, while a fundamental challenge remains: Can image generation be controlled at the element level, enabling intuitive modifications such as adjusting shapes, altering colors, or adding and removing objects? In this work, we address this challenge by introducing layer-wise controllable generation through simplified vector graphics (VGs). Our approach first efficiently parses images into hierarchical VG representations that are semantic-aligned and structurally coherent. Building on this representation, we design a novel image synthesis framework guided by VGs, allowing users to freely modify elements and seamlessly translate these edits into photorealistic outputs. By leveraging the structural and semantic features of VGs in conjunction with noise prediction, our method provides precise control over geometry, color, and object semantics. Extensive experiments demonstrate the effectiveness of our approach in diverse applications, including image editing, object-level manipulation, and fine-grained content creation, establishing a new paradigm for controllable image generation. 
Project page: \url{https://guolanqing.github.io/Vec2Pix/}
\end{abstract}

\section{Introduction}
Recent progress in image generation~\citep{rombach2022latent, ho2020denoising, peebles2023scalable, flux2024}, 
% exemplified by models such as GPT-4o~\citep{gpt4o2024} and Gemini-Flash-Image~\citep{gemini2024},
has led to remarkable advances in visual quality. 
However, in real-world scenarios, the results often fail to fully meet user expectations: local details may be unsatisfactory, and fine-grained controllability is lacking, making it difficult to flexibly edit specific regions or manipulate individual elements. 
Tasks such as adjusting an object’s shape, changing its color, or adding or removing elements usually require complex prompt engineering or specialized editing pipelines, which can introduce artifacts and disrupt background consistency.

To address controllability, prior works explore conditional guidance such as sketches~\citep{chen2009sketch2photo, wang2021sketch}, layouts~\citep{zhao2020layout2image, sun2021learning}, or drag-based interactions~\cite{pan2023drag}. 
These methods have benefited digital art and design, yet their operability remains limited --- controls are either too coarse (e.g., layout) or too localized (e.g., drag), and often lack semantic awareness of the edited elements thus not easy to edit shape and attributes. 
% While these methods have advanced digital art and design, their operability remains limited: controls are often either too coarse (e.g., layout) or too localized (e.g., drag), and they typically lack semantic awareness of the edited elements, making it difficult to modify shapes and attributes effectively.
This gap raises a fundamental question: \textit{can image generation be controlled at the element level, enabling intuitive modifications such as adjusting shapes or sizes, altering colors, or adding and removing objects?} 

In this work, we address this challenge by introducing a novel element-wise controllable generation framework by involving semantically-aligned vector graphics (VGs) as representations. Our key idea is to parse images into hierarchical vector representations, where the reconstructed vector graphics are both semantically aligned and structurally coherent. 
To this end, we propose an efficient vectorization pipeline that achieves over a $7\times$ speedup while producing VGs that are easy to modify. Building on this representation, we develop a generation framework guided by VGs, which enables users to freely modify elements and seamlessly translate these edits into photorealistic outputs. In particular, a tunable encoder predicts the initial noise from VGs, further aligning structural and semantic features for controllable synthesis.
Extensive experiments demonstrate that our approach enables flexible and reliable element-wise editing across diverse applications, including image editing, object-level manipulation, realistic and vector graphics composition, and localized de-artifacts. We believe this establishes a new paradigm for controllable image generation, bridging the gap between high-quality synthesis and practical usability. Our contributions can be summarized as:
% By leveraging the structural and semantic features of vector graphics in conjunction with noise prediction, our method provides precise control over geometry, color, and object semantics.

\begin{itemize}
    \item We propose a controllable generation framework that leverages \textit{Simplified Vector Graphics}, a hierarchical vector parsing of images that is semantically aligned and structurally coherent, offering interpretable and manipulable representations for user interaction.
    \item We present \textit{Vector-Guided Noise Prediction}, where a tunable encoder derives the spatially variant initial noise of diffusion models from vector graphics, thereby aligning structural and semantic features for controllable synthesis.
    \item Experimental results show that our framework supports element-wise re-generation and modification, and significantly improves controllable image generation, enabling layout-wise generation, object-level editing, and composition.
\end{itemize}

\section{Preliminary}

\noindent\textbf{Flow models}~\citep{liu2022flow,lipman2022flow,bortoli2022riemannian} parameterize the velocity field $u_t \in \mathbb{R}^d$. 
Recent advances such as Flux~\citep{flux2024} apply this principle to text-to-image generation, where an image $I$ is encoded into a latent variable $\bm{z}_0=\text{Enc}(I)$ by a Variational autoencoder (VAE) encoder $\text{Enc}(\cdot)$
consisting of convolutional downsampling layers, residual blocks, and a mid-level attention mechanism. 
A symmetric decoder reconstructs an image $\hat{I}$ from a latent representation $\hat{\bm{z}}_0$. The latent dynamics are modeled by a transformer-based velocity network, adopting a DiT~\citep{peebles2023scalable}-style design where latent patches are treated as tokens.

The flow model construction assumes a Gaussian prior $\bm{z}_1 \sim \mathcal{N}(0,I)$ and connects it to the data latent $\bm{z}_0$ through a linear interpolation path:
\begin{equation}
\bm{z}_t = (1-t)\bm{z}_1 + t\bm{z}_0 , \quad t \in [0,1],
\end{equation}
with instantaneous velocity $u_t(\bm{z}_t) = \bm{z}_0 - \bm{z}_1$.

The transformer denoiser $v_\theta$ is trained to approximate this velocity by minimizing the squared error:
\begin{equation}\label{eq:fm_loss}
\mathcal{L}_{\text{FM}}(\theta) = \mathbb{E}_{\bm{x}_0, \bm{z}_0, \bm{z}_1, t}\left[| v_\theta(\bm{z}_t, t, c) - (\bm{z}_0 - \bm{z}_1) |_2^2\right].
\end{equation}
Sampling starts from Gaussian noise $\bm{z}_1$ by solving the ODE
$\frac{d\bm{z}_t}{dt} = v_\theta(\bm{z}_t,t,c)$ from $t:1 \to 0$.
% \frac{d\bm{z}_t}{dt} = v_\theta(\bm{z}_t,t,c), \quad t:1 \to 0.
% \end{equation}
% using standard numerical solvers such as Euler or Heun. The resulting latent $\hat{\bm{z}}_0$ is then decoded into $\hat{\bm{x}}_0$ by the VAE decoder. 
% This construction provides a minimal yet functional connection between flow matching and latent-variable generative modeling, serving as a basis for further architectural and algorithmic refinement.
\begin{figure*}[t]
  \centering
  \vspace{-3pt}
  \includegraphics[width=.95\textwidth]{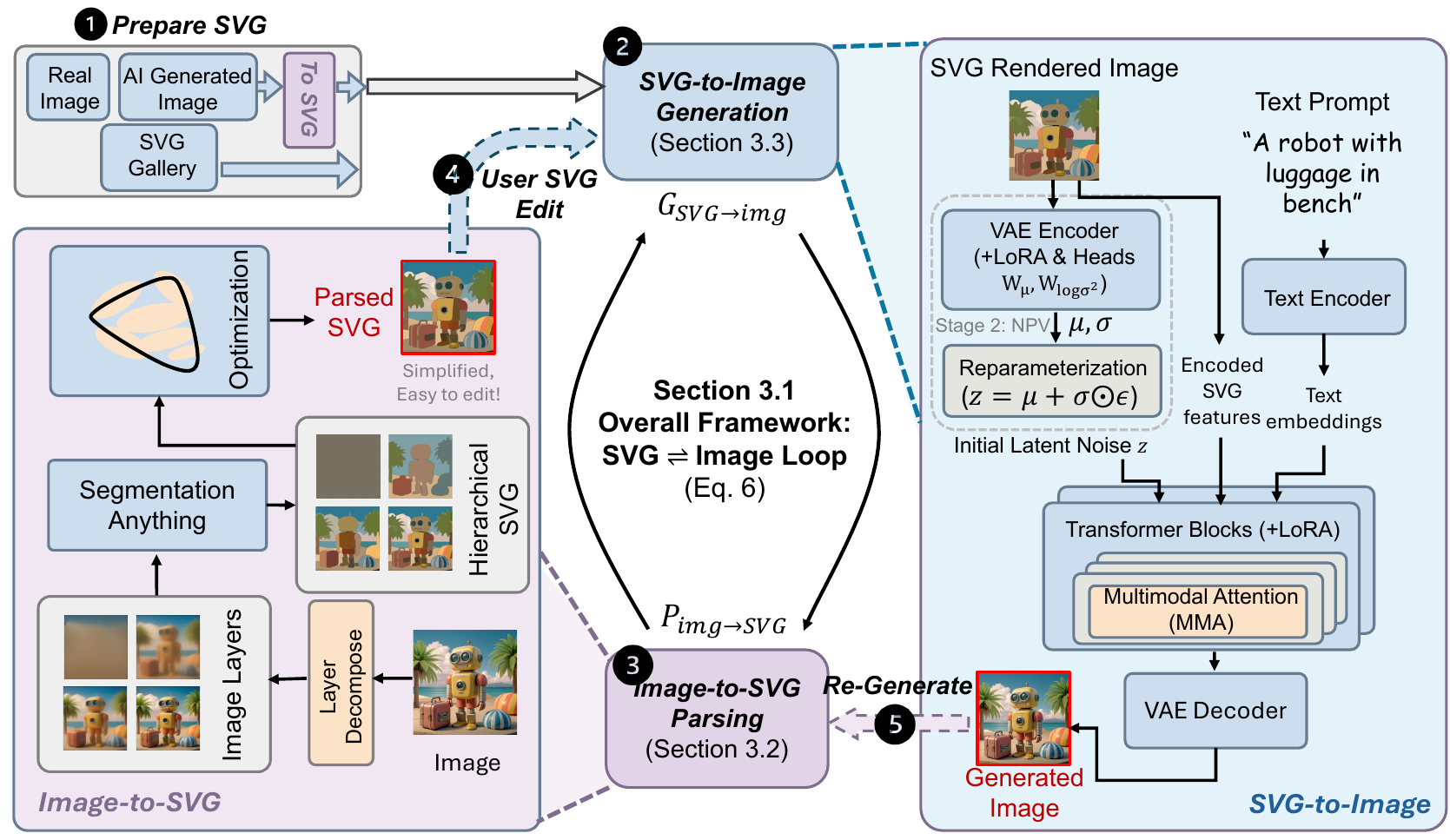}  % 或 .png/.jpg
  % \vspace{-6pt}
  \caption{Overall framework of our Vec2Pix and its workflow. 
\textcircled{\small 1} \textbf{Prepare SVG:} the input is obtained by converting a real or AI-generated image into SVG, or by selecting an existing SVG from a gallery. 
\textcircled{\small 2} \textbf{SVG-to-Image:} the SVG information will be conditioned using token concatenation and noise prediction from vectors (NPV) module. The NPV module incorporates the SVG condition and integrates trainable LoRA adapters and prediction heads to estimate the mean and variance of the initial noise, rather than directly sampling from Gaussian noise.
\textit{If the user wishes to re-generate or modify specific parts, we proceed with steps \textcircled{\small 3}–\textcircled{\small 5}.} 
\textcircled{\small 3} \textbf{Image-to-SVG:} the generated image is converted back into SVG using a diffusion model to produce multiple layers, followed by SAM to generate semantic masks for each layer, and further refined via 2D Gaussian optimization. 
\textcircled{\small 4} \textbf{SVG Editing:} users can interactively edit the SVG by adjusting curves and attributes. 
\textcircled{\small 5} \textbf{Re-generation:} the modified SVG is used as guidance to synthesize the final updated result.
}
\label{fig:method}
\end{figure*}

\noindent\textbf{Vector graphics (VG)} represent images not as dense pixels but as geometric primitives such as points, lines, and curves. 
% Unlike raster images, VG are resolution-independent and naturally support object-level editing. 
% Among different primitives, \emph{Bézier curves} are particularly popular for their ability to model smooth boundaries.  
% Most existing image vectorization methods represent an image as a collection of \emph{closed regions}, 
% each described by an ordered sequence of Bézier curve segments:
% \begin{equation}
% \Omega = \{ B_{i_1}, \dots, B_{i_L} \}, \; 
% B_{i_\ell}(1) = B_{i_{\ell+1}}(0), \; B_{i_L}(1) = B_{i_1}(0),
% \end{equation}
% where $\Omega$ denotes the closed region, and each $B_i$ is a Bézier segment. 
% The segments are ordered so that the endpoint of $B_{i_\ell}$ matches the start of $B_{i_{\ell+1}}$, 
% and the last segment $B_{i_L}$ connects back to the first $B_{i_1}$. 
% Thus, the index $i$ specifies the sequence of segments, ensuring that all $B_i$ are linked head-to-tail to form a continuous closed curve.

% A Bézier segment $B_i(t)$ of degree $M$ is parameterized as
% \begin{equation}
% \vspace{-1mm}
% B_i(t) = \sum_{j=0}^{M} \binom{M}{j} (1-t)^{M-j} t^j P^{(i)}_j, \; t \in [0,1], \; P^{(i)}_j \in \mathbb{R}^2.
% \vspace{-1mm}
% \end{equation}

% Here, \(M\) denotes the polynomial degree, and \(P^{(i)}_j\) are the control points. 
% In our paper, we used \emph{cubic Bézier curves} (\(M=3\)), as they strike a balance between expressiveness and simplicity.  

Among different vector primitives, Bézier curves are widely used for representing smooth boundaries.
Most image vectorization methods model an image as a collection of closed regions,
each defined by an ordered sequence of Bézier segments $\Omega=\{B_{i_1},\dots,B_{i_L}\}$ forming a continuous closed curve.
Each Bézier segment $B_i(t)$ of degree $M$ is parameterized as
\begin{equation}
B_i(t)=\sum_{j=0}^{M}\binom{M}{j}(1-t)^{M-j}t^j P^{(i)}_j,
\quad t\in[0,1],\;
\end{equation}
where $P^{(i)}_j\in\mathbb{R}^2$ and we use cubic Bézier curves ($M=3$) in all experiments.

The emergence of \emph{differentiable vector graphics rendering} (DiffVG~\citep{li2020differentiable} and Bézier Splatting \citep{liu2025b}) makes it possible to propagate gradients through rasterization. This allows the control points to be treated as \emph{optimizable parameters}, refined by minimizing the discrepancy between the rasterized result and the target image. A common objective is the pixel-wise reconstruction loss:
\begin{equation}
\mathcal{L}_{\text{recon}} = \big\| I_{\text{real}} - R(\{ \Omega_k \}) \big\|_1,
\label{eq:recon_loss}
\end{equation}
where \(R\) denotes the differentiable rasterizer applied to the set of regions \(\{ \Omega_k \}\). This formulation establishes a direct link between continuous vector representations and gradient-based optimization, providing the foundation for trainable and editable vector graphics.

\section{Methodology}

\subsection{Overall Framework}
Unlike prior controllable image generation models based on sketches, layouts, or depth maps~\citep{wang2021sketch,zhao2020layout2image, tan2025ominicontrol}, this work adopts \textbf{vector graphics}, like SVG, as the conditioning representation, offering a more flexible, structured, and easy-to-edit description of visual content. Vector graphics inherently capture the hierarchical relationships between foreground objects and background elements within a scene, enabling richer control and composition. This structure makes them more intuitive for user manipulation, enabling direct editing and control over individual components. In addition, vector graphics seamlessly integrate with existing SVG libraries, allowing easy reuse and composition of existing rich design assets.

% problem, why vector graphics

To this end, we design a loop of \textbf{SVG$\rightleftharpoons$image} consisting of two key components:
(1) an \textit{SVG-guided image generation module} that ensures the synthesized image faithfully follows both user-manipulated vector graphics and the accompanying text prompt (see Section~\ref{sec:vector}),
and (2) an \textit{image-to-SVG parsing module} that converts images back into vector graphics with hierarchical semantic representations (see Section~\ref{sec:generation}).
Together, these components form a closed loop that enables element-wise controllable generation and iterative refinement through user interactions.
Formally, the loop can be expressed as
\begin{equation}
\mathcal{S} 
\xrightarrow{\; G_{\text{SVG}\to\text{Img}} \;} \hat{\mathcal{I}} 
\xrightarrow{\; P_{\text{Img}\to\text{SVG}} \;} \hat{\mathcal{S}} 
\xrightarrow{\; \text{edit} \;} \hat{\mathcal{S}}' 
\xrightarrow{\; G_{\text{SVG}\to\text{Img}} \;} \hat{\mathcal{I}}',
\end{equation}
where $\mathcal{S}$ denotes the input SVG representation, $\hat{\mathcal{I}}$ is the generated image, $\hat{\mathcal{S}}$ is the reconstructed SVG, $\hat{\mathcal{S}}'$ is the user-modified SVG, and $\hat{\mathcal{I}}'$ is the re-generated image.

\subsection{Efficient and Simplified Vector Graphics}\label{sec:vector}
\begin{figure}[t] % r=右侧放图；宽度建议 0.35--0.5\textwidth
  % \vspace{-6pt}                        % 细调：让图片更贴近上方文字
  \centering
  % \includegraphics[width=0.9\linewidth]
  % {fig/artifact.pdf} % 换成你的图片文件
  \includegraphics[width=0.9\linewidth]
  {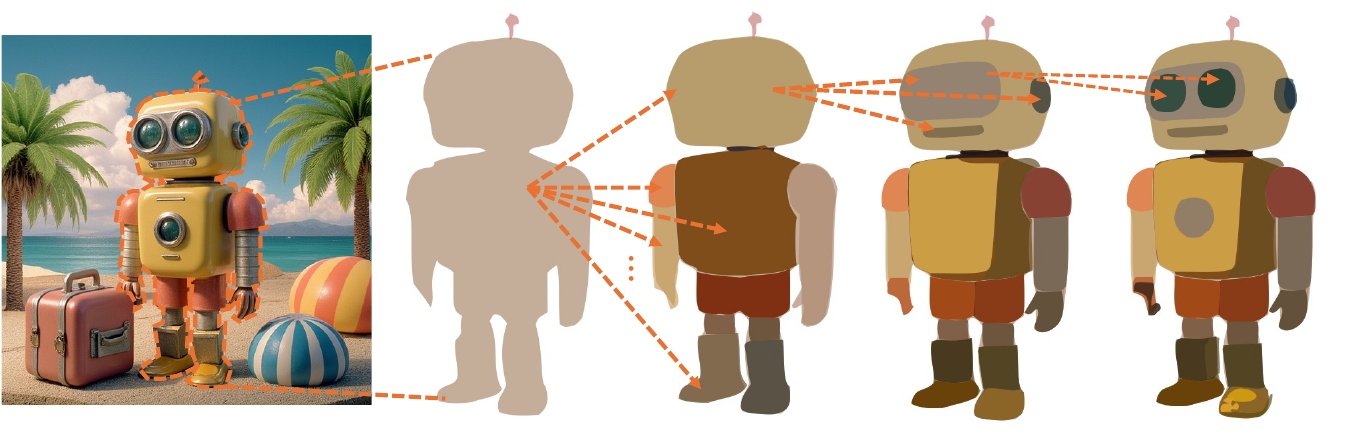} % 换成你的图片文件
  \vspace{-6pt}
  \caption{Our hierarchical and simplified vector graphics.
As illustrated, the SVG is decomposed into the robot ``body", which further contains semantic parts such as the ``head", ``upper body", ``legs", and ``shoes".
The ``head" is hierarchically subdivided into the ``eye region", ``ears", and ``mouth",
and the ``eye region" is further decomposed into the ``left eye" and ``right eye".}
  \label{fig:hierarchical}
  % \vspace{-10pt}                       % 细调：减小图后空白
\end{figure}

Unlike existing image vectorization approaches~\citep{illustrator2025, li2020differentiable, ma2022towards}, which primarily focus on reconstruction quality and often produce overly complex and fragmented SVGs without semantic alignment, existing designs lead to limited editability and low efficiency, making such representations unsuitable as conditioning inputs for image generation. To address this issue, we propose an Image-to-SVG parsing module that efficiently delivers an \textit{structurally simple, semantically aligned, and hierarchical} vectorized representation, which allows users to easily identify and manipulate complete objects that match their intent, and select an object together with all of its sub-parts in a unified and intuitive manner. Figure~\ref{fig:hierarchical} illustrates an example of hierarchical and simplified vector graphics generated by our method.

\noindent\textbf{Hierarchical and semantic-aligned initialization.}
We leverage visual foundation models, including Segment Anything~\citep{kirillov2023segment} and diffusion model ~\citep{rombach2022latent}, to initialize hierarchical and semantically aligned vector graphics, inspired by prior layer decomposition methods such as LIVSS~\citep{wang2025layered}. 
Specifically, diffusion priors are used to generate a sequence of progressively simplified images
$\{I_S^{t}\}_{t=1}^{n}$, where smaller $t$ captures global structure and larger $t$ preserves finer details.
Each $I_S^{t}$ is segmented by Segment Anything to obtain semantic masks, which are then assigned to layers based on simplification level, mask size, and overlap with previously assigned masks.
Redundant masks with significant occlusion are discarded, resulting in a hierarchical mask collection
\begin{equation}
\mathcal{M}
=
\left\{
\mathcal{M}_k
\right\}_{k=1}^{K},
\qquad
\mathcal{M}_k
=
\left\{
I_{\text{mask}}^{k,j}
\right\}_{j \in \mathcal{J}_k},
\end{equation}
where $k$ indexes the hierarchy level and $|\mathcal{J}_k|$ may vary across layers.
Each mask is associated with a parent mask in the preceding layer.
 % mask boundaries are polygonized and simplified using the Douglas--Peucker algorithm~\citep{Douglas1973ALGORITHMSFT}. 
Building upon the hierarchical masks, we initialize each semantic region by polygonizing its boundary and converting it into an editable SVG path: we simplify the polygon (Douglas--Peucker), and if it remains overly complex, we split it at the longest diagonal and fit each sub-chain with a few cubic Bézier segments, limited at 8 segments per side to ensure the edibility.

\noindent\textbf{Efficient image vectorization via differentiable rasterization.}
To enable fast image vectorization within image-to-SVG loop, we adopt Bézier Splatting~\cite{liu2025b} as an efficient differentiable renderer. However, directly adopting Bézier Splatting introduces a challenge in preserving the initial hierarchical semantic alignment.
the standard alpha blending formulation permits opacity manipulation, which may corrupt the exclusivity of semantic regions
and break the hierarchical structure during optimization.

To address this issue, we enforce a semantic-aligned rendering formulation in which each vector graphic exclusively belongs to a single semantic region within a layer.
Since semantic regions are not additive under alpha blending, we fix the opacity of all regions to $1.0$,
ensuring that optimization preserves hierarchical semantic alignment through boundary adjustment rather than opacity manipulation.

Under this formulation, we optimize the hierarchically initialized vector graphics $\{\Omega_k\}$ with explicit structural supervision.
Specifically, we introduce a structure loss that matches each rendered SVG region to its corresponding semantic mask.
The overall optimization objective is
\begin{equation}
\mathcal{L}(\{\Omega_k\})
=
\sum_{k=1}^{K}
\left\|
I_{\text{mask}}^k - I_{\text{vec}}^k
\right\|
+
\gamma \, \mathcal{L}_{\text{recon}},
\end{equation}
where $I_{\text{vec}}^k$ denotes the SVG rendering at layer $k$,
$\mathcal{L}_{\text{recon}}$ is the image-level reconstruction loss defined in Eq.~\ref{eq:recon_loss},
and $\gamma$ balances structural alignment and photometric fidelity.
This objective enforces semantic boundary consistency while maintaining accurate appearance reconstruction.

\subsection{Vector-Guided Controllable Image Generation}\label{sec:generation}
% \noindent\textbf{Motivation.} 
Since our SVG representation preserves richer and more accurate structural information than prior conditions such as sketches or depth maps, we introduce a dedicated Noise Prediction from Vector (NPV) module along with a two-stage training strategy within the SVG-to-Image module. This design better aligns SVG conditions with the generated images, particularly with respect to color fidelity and boundary geometry. Inspired by prior training-free image editing methods that leverage initial noise rescheduling \cite{rout2024semantic,mengsdedit} to maintain fidelity during editing, we move beyond fixed, hand-crafted noise formulations. Specifically, instead of computing the initial noise using a static training-free scheme, we propose an NPV module that learns to predict the initial noise distribution, including both mean and variance, conditioned on the rendered SVG images.
% Since our SVG representation preserves richer and more accurate structural information than prior conditions (e.g., sketch or depth), we introduce a dedicated Noise Prediction from Vector (NPV) module and two-stage training strategy in SVG-to-Image module to better align the SVG conditions with the generated results, particularly in terms of color fidelity and boundary geometry. Inspired by previous training-free image editing approaches, employing initial noise rescheduling~\cite{rout2024semantic,mengsdedit} to preserve fidelity while editing. Instead of computing the initial noise using a fixed training-free formulation, we introduce a NPV module that learns a predictable initial noise with both mean and variance conditioned on the SVG rendered images. 
This allows the model to more effectively inject conditional information while preserving high fidelity in the generated results.

In details, we introduce a \textbf{two-stage training strategy} for the SVG-guided image generation process $G_{\text{SVG}\to\text{Img}}$.  
In the \textbf{first stage}, we integrate LoRA modules into transformer blocks to adapt the original text-to-image model, \textit{i.e.}, Flux.1-dev~\citep{flux2024}, to text and SVG conditioned generation model.
The image branch encodes the parsed SVG into feature representations, which are concatenated with noise along the channel dimension to initialize generation. In parallel, the text branch encodes the accompanying description to provide semantic guidance. These two branches are fused through multimodal attention, allowing the model to jointly attend to visual and textual cues~\citep{tan2025ominicontrol}. We formulate this integration as:
\begin{equation}
Q = [Q_t; Q_z; Q_c], \; 
K = [K_t; K_z; K_c], \; 
V = [V_t; V_z; V_c],
\end{equation}
\begin{equation}
\mathrm{MMA}(Q,K,V) 
= \mathrm{softmax}\!\left(\frac{QK^{T}}{\sqrt{d}}\right)V.
\end{equation}
where $[;]$ represents the concatenation operation, and $Q$, $K$,
and $V$ are the query, key, and value components of the attention mechanism.

We then propose \textbf{Noise Prediction from Vectors (NPV)} as the \textbf{second stage} to further strengthen the structural alignment between SVGs and generated images. 
Specifically, we add a noise prediction module that estimates the mean and variance of the initial noise of flow matching model, such that the guidance from SVGs is directly injected at the beginning phase of the generation process.
We extend the Flux model's VAE encoder with low-rank adapters (LoRA) and lightweight head tuning to obtain the latent distribution parameters. Given an input image $x$, the encoder produces the mean $\mu$ and the log-variance $\log\sigma^2$ as
\begin{equation}
\mu, \log \sigma^2 
= \big(W_\mu, W_{\log\sigma^2}\big) * \phi\!\left(\text{Enc}_{\widetilde{\theta}}(I)\right), 
\end{equation}
where
$\text{Enc}_{\widetilde{\theta}}(\cdot)$ denotes the frozen Flux VAE encoder plugged with learnable LoRA modules, where the parameters are modified as $\widetilde{\theta} = \theta^{(0)} \cup \{\Delta W_l\}_l$ with LoRA updates injected into selected layers. Here, $\phi$ represents a GroupNorm layer followed by a SiLU activation, and $W_\mu, W_{\log\sigma^2}$ are parallel $1\times1$ convolutional heads predicting $\mu$ and $\log\sigma^2$, respectively.
% \begin{itemize}
%     \item $\text{Enc}_{\text{Flux}}(x)$: output of the frozen Flux VAE encoder, 
%     \item $\text{LoRA}(x)$: low-rank residual adapters injected into selected layers,
%     \item $\phi$: GroupNorm followed by SiLU activation, 
%     \item $W_\mu, W_{\log\sigma^2}$: parallel $1\times1$ convolution heads predicting $\mu$ and $\log\sigma^2$,
%     % \item $\log\sigma^2$ (often denoted as $\logvar$): the logarithm of the variance, \textit{i.e.}, $\log\sigma^2$.
% \end{itemize}

With these parameters, we sample the latent code $z$ via the reparameterization trick:
\begin{equation}
z = \mu + \sigma \odot \epsilon, 
\quad \epsilon \sim \mathcal{N}(0,I).
\end{equation}
The training objective consists of three components: the flow matching loss $\mathcal{L}_{\text{FM}}$ from Eq. \ref{eq:fm_loss},
the KL loss $\mathcal{L}_{\text{KL}}$ to minimize the divergence from the standard normal distribution, and a covariance loss $\mathcal{L}_{\text{cov}}$ to encourage spatially independence across latent channels:  
\begin{equation}
\mathcal{L}_{\text{KL}} = 
\frac{1}{2}\,\mathbb{E}\Big[\mu^2 + \sigma^2 - \log\sigma^2 - 1\Big].
\end{equation}
\begin{equation}
\mathcal{L}_{\text{cov}} = \frac{1}{C(C-1)} 
\sum_{i \neq j} R_{ij}^2, 
\qquad R = \frac{\tilde{\mu}\,\tilde{\mu}^\top}{\|\tilde{\mu}\|^2},
\end{equation}
where $\tilde{\mu}$ is the channel-normalized mean vector, and $C$ is the number of elements in the window. To avoid the large memory overhead and artifacts caused by computing covariance with a global fixed window, we randomly select $N$ patches of size $p\times p$ and compute the mean covariance loss over the patches.
%
% \begin{itemize}
%     \item \textbf{KL Loss}: KL divergence between the approximate posterior $q(z|x)=\mathcal{N}(\mu,\sigma^2 I)$ and the standard normal prior $p(z)=\mathcal{N}(0,I)$:
%     \begin{equation}
% \mathcal{L}_{\text{KL}} = 
% \frac{1}{2}\,\mathbb{E}\Big[\mu^2 + \sigma^2 - \log\sigma^2 - 1\Big],
% \end{equation}
% \item \textbf{Covariance Loss}: 
% to encourage independence across latent channels:
% \begin{equation}
% \mathcal{L}_{\text{cov}} = \frac{1}{C(C-1)} 
% \sum_{i \neq j} R_{ij}^2, 
% \qquad R = \frac{\tilde{\mu}\,\tilde{\mu}^\top}{\|\tilde{\mu}\|^2},
% \end{equation}
% where $\tilde{\mu}$ is the channel-normalized mean vector.
% \end{itemize}
The final objective is a weighted combination, as:
\begin{equation}
\mathcal{L} = \mathcal{L}_{\text{FM}} + \beta \,\mathcal{L}_{\text{KL}} + \lambda \,\mathcal{L}_{\text{cov}}.
\end{equation}

\begin{figure*}[t]
  \centering
  % \vspace{-15pt}  
  \includegraphics[width=\textwidth]{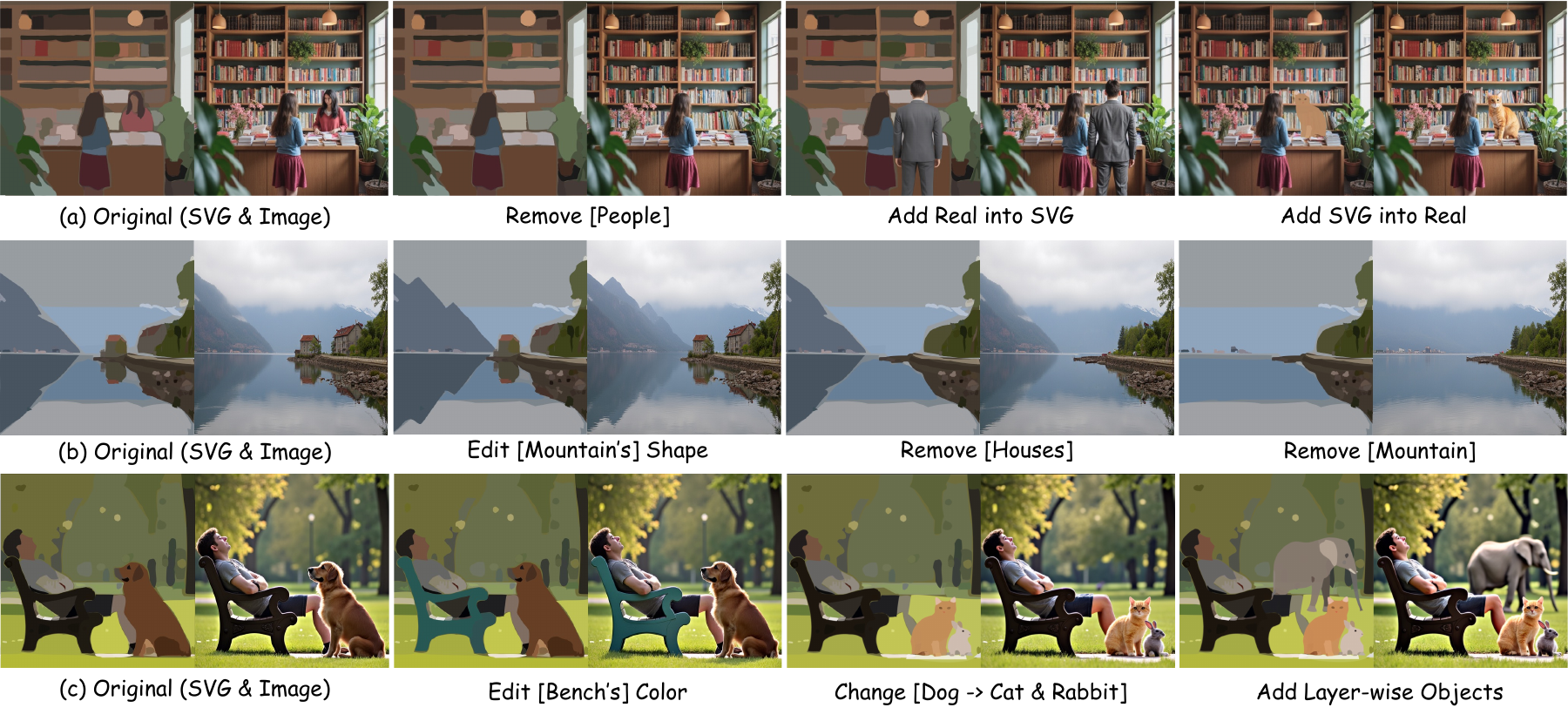}  % 或 .png/.jpg
  \vspace{-16pt}
  \caption{Our Vec2Pix supports various controllable image generation and editing tasks. 
}
% \vspace{-20pt}
  \label{fig:examples}
\end{figure*}

\vspace{-10pt}
\section{Experiments}

\subsection{Experimental Setups}

\noindent\textbf{Implementation details.} Our Vec2Pix builds on FLUX.1-dev~\citep{flux2024}, a latent rectified flow transformer for text-to-image generation. We apply LoRA~\citep{hu2022lora} to the transformer blocks of the base model, using a default rank of $4$, with the LoRA scale set to $0$ by default for non-condition tokens, introducing approximately 14.5M trainable parameters for stage one. In addition, we apply LoRA with rank of $8$ to the Flux VAE encoder at both the down and middle stages, introducing approximately 135K trainable parameters to predict the initial noise.
Our model is trained with a batch size of $1$ and gradient accumulation over $4$ steps. We employ the Prodigy optimizer with
safeguard warmup and bias correction enabled, setting the
weight decay to $0.01$. We set hyperparameter $\gamma=\beta=\lambda=1$ and the number of sampled patches as $N=8192$ and patch size $p=4$.
All experiments are conducted on 4 NVIDIA A100 GPUs (80GB each).
For stage one, our model is trained for 50K iterations, while stage two is trained for 10K iterations.

\noindent\textbf{Dataset construction.} 
We filter around 5M images and their corresponding text prompts from the LAION-400M dataset, retaining only those with resolutions larger than $1024 \times 1024$. All training data are resized to $512 \times 512$ for training. We then employ our proposed efficient Image-to-SVG module to convert these images into SVGs, thereby constructing \{image, SVG, text\} triplets for the training set.
We adopt this dataset consistently for both stage one and stage two training processes.

% \vspace{-10pt}
\subsection{Applications}

\begin{figure*}[t]
  \centering
  \vspace{-5pt}
  \includegraphics[width=\textwidth]{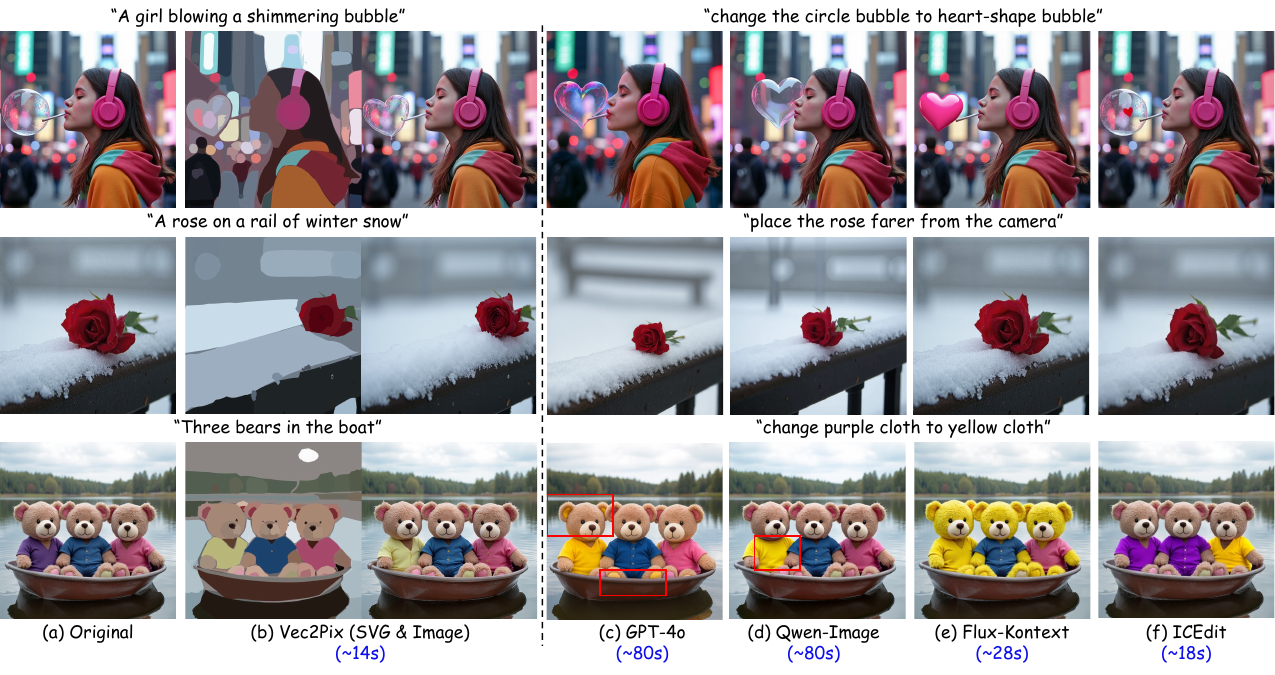}
  \vspace{-20pt}
  \caption{Visual comparisons with text-prompt-guided editing methods, including state-of-the-art open-source and commercial solutions such as GPT-4o~\cite{gpt4o2024}, Qwen-Image~\cite{wu2025qwen}, Flux-Kontext~\cite{labs2025flux}, and ICEdit~\cite{zhang2025context}. Editing cases such as shape modification, object repositioning, and color adjustment are readily supported by our VG representation, whereas text-guided editing often fails.
}
  \label{fig:compare_edit}
\end{figure*}

\begin{figure*}[t]
  \centering
   % \vspace{-16pt}
  \includegraphics[width=\textwidth]{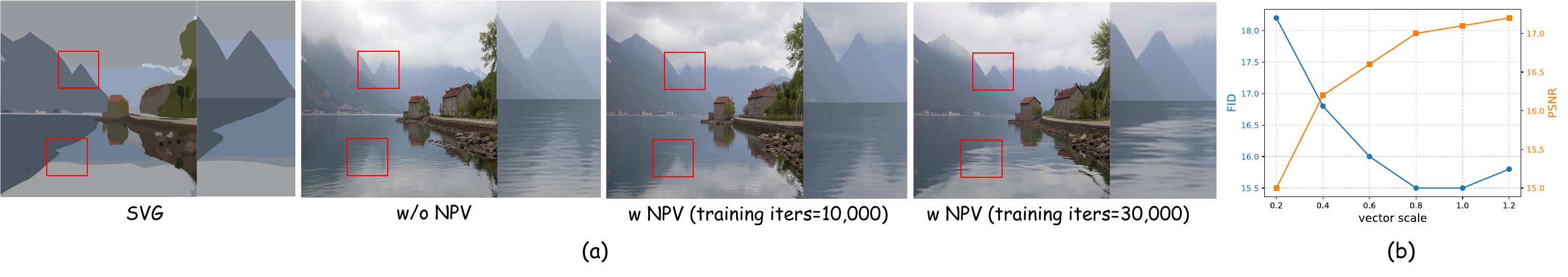}
  \vspace{-20pt}
  \caption{(a) Ablation study comparing results with and without the proposed Noise Prediction from Vectors (NPV) module, as well as training NPV module with different iterations.
(b) PSNR and FID performance variations under different vector scales used to adjust the conditioning strength. By adjusting the vector scale, our method can flexibly adapt to complex appearance effects such as water reflections, smoke, and lighting variations, without being overly constrained by the SVG geometry.
}
  \label{fig:ablation}
\end{figure*}

% As shown in Figure \ref{fig:examples}, our Vec2Pix method supports diverse re-generation and editing tasks, including:

Figure~\ref{fig:examples} showcases the applications supported by our Vec2Pix method, while Figure~\ref{fig:compare_edit} compares its editing performance with recent state-of-the-art approaches. Vec2Pix enables a wide range of re-generation and editing tasks, including:

\begin{figure}[t] % r=右侧放图；宽度建议 0.35--0.5\textwidth
  % \vspace{-6pt}                        % 细调：让图片更贴近上方文字
  \centering
  % \includegraphics[width=0.9\linewidth]
  % {fig/artifact.pdf} % 换成你的图片文件
  \includegraphics[width=0.9\linewidth]
  {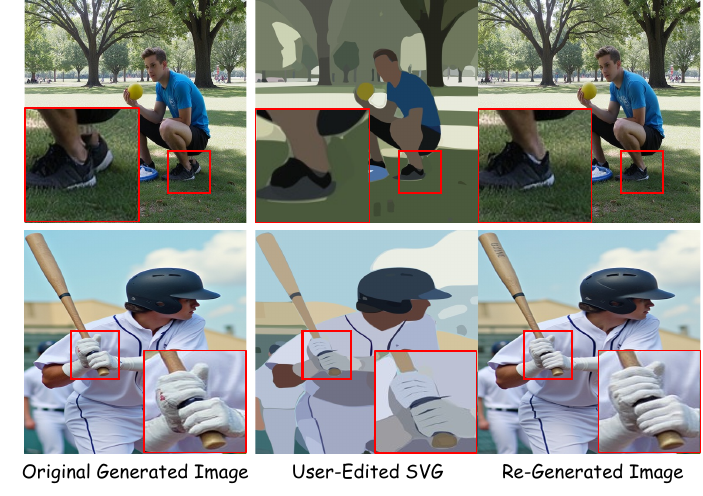} % 换成你的图片文件
  \vspace{-6pt}
  \caption{Our Vec2Pix can be applied to artifact removal. Fine-grained elements in the original generation, such as shoes and fingers, are sometimes rendered incorrectly (\textit{e.g.}, \textbf{distorted shoes} and \textbf{three fingers} instead of four). By applying manual corrections or guidance in SVG, the re-generated image can be effectively refined to eliminate such artifacts.}
  \label{fig:artifact}
  % \vspace{-10pt}                       % 细调：减小图后空白
\end{figure}

\noindent\textbf{Layer-wise generation.}
Vec2Pix enables controllable layer-wise generation, where different semantic layers (\textit{e.g.}, background, mid-level structures, and foreground details) can be synthesized independently. For example, the object can be positioned at any layer, such as placing the building behind existing trees, inserting furniture within the mid-level interior space.
This design enables flexible scene composition while keeping local edits semantically consistent and visually faithful, through SVG that preserves semantic features via hierarchical layering.
% This design provides flexible scene composition while ensuring that local modifications remain consistent with global semantics and preserve visual fidelity. 
% Such capability arises from our SVG parsing strategy, which retains semantic features with explicit hierarchical relationships, allowing elements to be composited in a structured, layer-by-layer manner.
% one can first generate a clean background scene, then refine architectural structures, and finally add fine-grained textures such as vegetation or human figures. This hierarchical control facilitates targeted refinements while maintaining overall coherence and fidelity.

\noindent\textbf{Object editing.}
Vec2Pix supports object editing, allowing precise manipulation of individual components within a scene. Users can alter object attributes such as shape, color, or position, which is especially beneficial for design and creative applications. For instance, replacing a dog with a cat and rabbit, changing the color of a bench, or modifying the shape of mountain. These edits preserve contextual consistency, keeping the scene semantically coherent and visually realistic.

\noindent\textbf{Reference-based generation.}
Vec2Pix enables reference-guided synthesis, in which objects' structures and color can be transferred from one or multiple exemplars, in the mean time, the SVG controls structures. 
Then generated harmonized results in the scene such as the inserted cat, rabbit, and elephant in Figure~\ref{fig:examples}(b).
% References can be applied either globally or to user-defined masks, with tunable strength to trade off identity preservation and scene coherence. 
% Multiple references may be layered (\textit{e.g.}, style from artwork + materials from swatches) and combined with SVG constraints to target specific objects. 
Vec2Pix can mitigate background leakage and resolves conflicting cues via learned priors, yielding photorealistic results faithful to both the references and the SVG layout.

\noindent\textbf{Real \& SVG composition.}
By leveraging vector-based SVG representations, our approach enables flexible composition of SVGs and real images within a single complex scene. Vector primitives can be rearranged, combined, or modified to form realistic scenes such as the cat in a bookstore shown in Figure~\ref{fig:examples}(a), and also allow the insertion of realistic objects (e.g., the man) into SVG-based conditions.

% Vector primitives can be rearranged, combined, or modified, for example, layering shapes to form a logo, assembling furniture layouts in an interior design sketch, or merging geometric icons to create stylized illustrations. 
% The framework translates these structured manipulations into photorealistic images.
% , ensuring both structural alignment and semantic faithfulness. 

\begin{table*}[t]
\centering
\small
% \vspace{-10pt}
\caption{\blue{Quantitative comparison with controllable generation baselines across different representation types including sketch, depth, stroke, segmentation mask, and our proposed SVG, along with our variants trained with and without noise prediction from vectors (NPV). We report both reconstruction and editing quality, with the best results highlighted in \textbf{bold}.}}
\setlength{\tabcolsep}{5pt}
\adjustbox{width=.8\textwidth}{
\begin{tabular}{l|c|cccc|cccc}
\toprule
\multirow{2}{*}{\textbf{Representation}}  &\multirow{2}{*}{\makecell{\textbf{Base}\\\textbf{Model}}} &
\multicolumn{4}{c|}{\textbf{Reconstruction}} &
\multicolumn{4}{c}{\textbf{Editing}} \\
& & \textbf{FID$\downarrow$}& \textbf{LPIPS$\downarrow$}  & \textbf{SSIM$\uparrow$} & \textbf{PSNR$\uparrow$}& \textbf{FID$\downarrow$}& \textbf{LPIPS$\downarrow$}  & \textbf{SSIM$\uparrow$} & \textbf{PSNR$\uparrow$}  \\
\midrule
Canny &  FLUX.1 & 17.38 & 0.4582 & 0.42 & 12.28 & 22.48 & 0.5639 & 0.37 & 11.54 \\
% \blue{Canny w NPV} &  FLUX.1 & 16.89 & 0.4501 & 0.44 & 12.65 &21.76 & 0.5573 & 0.41 & 12.32 \\

Depth   &  FLUX.1         & 19.65 & 0.5572 & 0.36 & 11.60 & 22.94 & 0.5932 & 0.36 & 11.37 \\
Stroke   &  FLUX.1       & 22.42 & 0.4721 & 0.45 & 17.13 & 25.35 & 0.5249 & 0.44 & 14.87\\
\blue{Segmentation Mask} & FLUX.1 &  17.75 &0.4513 &0.43 & 15.78 & 21.97& 0.5382 &0.43&14.34\\
% \blue{Stroke w NPV} &  FLUX.1 & 21.88  & 0.4682 & 0.47  & 17.33  &25.01   & 0.5204 & 0.46 & 15.13 \\
Ours (SVG) w/o NPV  &  FLUX.1     & 16.03 & 0.3687 & 0.47 & 17.42 & 18.87 & 0.3901 & 0.47 & 17.25 \\
\rowcolor{rowhl}
Ours (SVG) w NPV  &  FLUX.1     & \best{15.52} & \best{0.3515} & \best{0.49} & \best{17.77} & \best{17.84} & \best{0.3836} & \best{0.50} & \best{17.72} \\
\bottomrule
\end{tabular}}
\label{tab:control-image-quality-alignment}
\end{table*}

\noindent\textbf{Localized de-artifacts.}
Vec2Pix can also address visual artifacts. When fine-grained details in the initial generation are flawed, such as fingers, are sometimes rendered incorrectly (\textit{e.g.}, wrong number of fingers),
manual adjustments or SVG-based guidance enable the re-generated image to be refined and corrected as shown in Figure~\ref{fig:artifact}.

\subsection{Method Comparison}

To evaluate the representation ability and controllability of Vec2Pix, we construct an evaluation set by collecting 5K paired source–target images from existing datasets~\citep{yu2025anyedit}. 
\blue{Each source image is converted into four structural conditions, including Canny edges, depth maps, strokes~\citep{mengsdedit}, Segmentation Mask~\cite{kirillov2023segment}, and our proposed SVG representation, using publicly available methods.} 
We then perform conditional generation to reconstruct the source image from each condition. 
\blue{Specifically, we adopt the well-trained Canny- and depth-conditioned models released in OmniControl~\citep{tan2025ominicontrol}, while the stroke and segmentation mask-conditioned model are trained with the same data as our SVG-conditioned model.} 
All methods, including ours, are built upon the FLUX.1-dev base model.
For evaluation, we measure reconstruction quality with FID~\citep{heusel2017gans}, 
PSNR~\citep{hore2010image}, 
SSIM~\citep{wang2004image}, 
and LPIPS~\citep{zhang2018unreasonable}.
 Beyond reconstruction quality, we assess controllability by examining the editability of different representations. Given source–target pairs, we compute modification masks and automatically apply them to the representations (\textit{e.g.}, $\text{mask} \times \text{target}_{\text{SVG}} + (1 - \text{mask}) \times \text{source}_{\text{SVG}}$). The edited representations are then used for conditional generation, and the resulting images are compared against the target images to quantify editing performance. Table~\ref{tab:control-image-quality-alignment} summarizes the reconstruction and editing results, demonstrating that our SVG serves as an effective representation for both information caching and editability.

%=== Table (放到文中需要的位置) =======================================

\subsection{Ablation Study}

\noindent\textbf{The effect of noise prediction.} In stage one, training treats SVGs as conditional inputs via token concatenation. Stage two adds noise prediction through the NPV module, further aligning SVG structure with generated outputs. As shown in Table~\ref{tab:control-image-quality-alignment} compares reconstruction quality with and without NPV variances, showing that incorporating NPV leads to better alignment in the generated results.
Besides, as shown in Figure~\ref{fig:ablation}(a), incorporating NPV makes the structure of the generated results align more closely with the input SVG, for example, the mountain silhouette.
Importantly, the mountain’s reflection in the generated image still obeys symmetry about the waterline, as encoded by the base model’s physics-consistent prior, despite a non-physical input SVG (\textit{e.g.}, asymmetric reflection edges). With more stage-two training iterations, adherence to the input structure strengthens, sometimes forcing physically implausible reflections.

\noindent\textbf{Effect of vector scale adjustment.} 
We adjust the strength of the SVG condition by varying a scaling factor, where we plot the PSNR and FID metrics under different scaling values, 
demonstrating how the factor influences both reconstruction quality and perceptual fidelity as shown in Figure~\ref{fig:ablation}(b).

\noindent\textbf{Effect of efficient and semantic-aligned vector graphics.}
To assess the effectiveness of our efficient and semantic-aligned image-to-SVG module, 
we compare it with the previous semantic-layered vectorization method LIVSS~\citep{wang2025layered} on the DIV2K dataset~\citep{agustsson2017ntire}, 
which contains 1K high-quality images. 
All methods are optimized for 30 steps and resized to $512 \times 512$ resolution for a fair comparison. 
Table~\ref{tab:efficient} reports results on both efficiency and reconstruction quality. 
The results demonstrate the advantages of our compact initialization and efficient Bézier Splatting rasterizer: 
under the same 30 optimization steps, our method achieves higher PSNR while being $7\times$ faster.

% oth LIVSS and our method are optimized for 30 steps on the DIV2K \citep{Agustsson_2017_CVPR_Workshops} dataset, with images resized to $512 \times 512$ resolution. 
% Reported time measures only the vector graphics optimization stage. 
% This ablation highlights the effect of our compact initialization and efficient Bézier Splatting rasterizer: under the same 30 optimiztion steps, our method achieves higher PSNR while being $7\times$ faster.

\begin{table}[t]
\centering
\caption{Comparison of reconstruction quality and efficiency between the existing semantic-layered image vectorization method LIVSS~\citep{wang2025layered} and our proposed efficient, semantically simplified image vectorization module.}
% \vspace{-5pt}
\adjustbox{width=.5\textwidth}{
\begin{tabular}{lcccc}
\toprule
\textbf{Method} & \textbf{PSNR $\uparrow$} & \textbf{Time (s) $\downarrow$} & \textbf{Speedup} \\
\midrule
LIVSS~\citep{wang2025layered} & 16.960 & 18.564 & 1.0$\times$ \\
\rowcolor{rowhl}
Our image vectorization                 & \textbf{17.353} & \textbf{2.656}  & \textbf{7.0$\times$} \\
\bottomrule
\end{tabular}
\vspace{-10pt}
}
\label{tab:efficient}
\end{table}

% traditional method;
% diffvg;
% ours;

% \vspace{-6pt}
\section{Related Work}
% \vspace{-6pt}

\noindent\textbf{Image vectorization.}
% Research on image vectorization has advanced rapidly in recent years. Traditional tools such as Image Trace in Adobe Illustrator can quickly convert raster images into vector graphics, but the pipeline is non-differentiable, often producing overly complex shapes that are difficult to edit. 
Image vectorization has advanced rapidly, but traditional tools like Illustrator’s Image Trace are non-differentiable, often producing overly complex shapes that are difficult to edit. 
A key breakthrough came with DiffVG \citep{li2020differentiable}, which pioneered differentiable rasterization and enabled gradient-based optimization of arbitrary Bézier curves, laying the foundation for compact and flexible vectorization. Building on this, LIVE \citep{ma2022towards} and O\&R \citep{hirschorn2024optimize} introduced layer-wise initialization strategies to generate more compact, topology-preserving representations, while LIVSS \citep{wang2025layered} further integrated SegmentAnything (SAM) \citep{kirillov2023segment} and diffusion priors to align SVGs with semantic content. Despite these advances, differentiable methods remain computationally expensive, often requiring hours for high-resolution optimization. Recently, Bézier Splatting \citep{liu2025b} recasts rasterization as splatting, yielding significantly faster optimization while preserving Bézier-curve flexibility. In this work, we integrate Bézier Splatting with layer-wise semantic alignment to enable high-quality image vectorization in seconds.

\noindent\textbf{Controllable image generation.}
Controllable generation in diffusion models has progressed from early text-to-image systems~\citep{rombach2022latent, saharia2022photorealistic} to spatially guided methods such as ControlNet~\citep{zhang2023adding}; UniControl~\citep{qin2023unicontrol} further unifies diverse spatial conditions under a Mixture-of-Experts (MoE) paradigm. However, because these approaches inject spatial condition features into the denoising hidden states, they are best suited to aligned inputs and struggle with misaligned or subject-driven generation. Extensions like IP-Adapter~\citep{ye2023ip} (cross-attention with an auxiliary encoder) strengthens identity preservation, yet most methods still operate in pixel space~\citep{qi2023fatezero,ruiz2023dreambooth,tumanyan2023plug,brooks2023instructpix2pix,kim2022diffusionclip,kawar2023imagic}.In contrast, we move beyond purely spatial conditioning and explore \emph{layer-wise controllable generation}, offering a more flexible and intuitive representation for fine-grained editing.
% Controllable generation has been extensively investigated in the context of diffusion models. 
% Early text-to-image frameworks~\citep{Rombach2022, Saharia2022} laid the foundation for conditional generation, while subsequent studies incorporated additional control modalities beyond text. 
% For example, ControlNet~\citep{Zhang2023ControlNet} enables spatially aligned conditioning through structural guidance, and T2I-Adapter~\citep{Mou2023Adapter} improves efficiency by injecting lightweight adapter modules. 
% UniControl~\citep{Zhao2023UniControl} further unifies diverse spatial conditions under a Mixture-of-Experts (MoE) paradigm, reducing the overall model footprint. 
% However, these approaches primarily operate by injecting spatial condition features into the hidden states of the denoising process, inherently constraining their applicability to spatially misaligned tasks such as subject-driven generation. 
% To overcome this limitation, IP-Adapter~\citep{Ye2023IPAdapter} introduces cross-attention with an additional encoder, while SSR-Encoder~\citep{Cao2023SSR} enhances identity preservation in image-conditioned scenarios. 
% Despite these advances~\citep{Qin2023, Ruiz2023, Shi2023, Tumanyan2023, Wang2023, Zhang2023InstructPix2Pix}, existing methods largely remain limited to pixel-space manipulations.
% In contrast, our work moves beyond spatial conditioning by exploring \emph{layer-wise controllable generation}, which provides a more flexible and intuitive representation for editing.
% \vspace{-7pt}
\section{Conclusion}
% \vspace{-5pt}
In this work, we have presented a new paradigm for controllable image generation based on semantic-aligned vector graphics. By decomposing images into hierarchical, semantically aligned vector representations and integrating them into a noise-guided generation framework, our method enables precise element-level control over geometry, color, and object semantics. The proposed approach not only delivers photorealistic outputs consistent with user edits but also supports a wide range of applications, from intuitive image editing to fine-grained object manipulation. These results highlight the potential of vector-guided generation as a foundation for the next generation of controllable and creative image synthesis.

% \subsubsection*{Author Contributions}
% If you'd like to, you may include  a section for author contributions as is done
% in many journals. This is optional and at the discretion of the authors.

% \subsubsection*{Acknowledgments}
% Use unnumbered third level headings for the acknowledgments. All
% acknowledgments, including those to funding agencies, go at the end of the paper.

% \section*{Broader Impact}
% This work aims to advance controllable image generation by introducing semantically aligned vector graphics as an interpretable and editable representation. The proposed method can benefit creative design, content creation, and human–computer interaction by enabling more intuitive and fine-grained image editing. As with other generative image models, potential downstream misuse may arise depending on application contexts, but these concerns are not unique to this work. We do not identify any specific ethical issues or negative societal impacts directly introduced by the proposed approach.

\bibliography{example_paper}
\bibliographystyle{icml2026}
\clearpage

\renewcommand{\thesubsection}{A.\arabic{subsection}}

\end{document}